\documentclass[a4paper,fleqn]{cas-dc}

\usepackage[numbers]{natbib}
\usepackage{graphicx}
\usepackage{amsmath,amssymb,amsfonts}
\usepackage{algorithmic}
\usepackage{textcomp}
\usepackage{subfigure} 
\newcommand{\etal}{\textit{et al.}\@ }
\newcommand{\ie}{\textit{i.e.}\@ }
\usepackage{multirow}

\def\tsc#1{\csdef{#1}{\textsc{\lowercase{#1}}\xspace}}
\tsc{WGM}
\tsc{QE}

\begin{document}
\let\WriteBookmarks\relax
\def\floatpagepagefraction{1}
\def\textpagefraction{.001}

\shorttitle{Generative Convolution Layer for Image Generation}    

\shortauthors{S. Park and Y.-G. Shin}  

\title [mode = title]{Generative Convolution Layer for Image Generation}  



%

\author[1]{Seung Park}[type=editor]

\ead{spark.cbnuh@gmail.com}

\affiliation[1]{organization={Biomedical Engineering, Chungbuk National University Hospital},
            addressline={776}, 
            city={Seowon-gu, Cheongju-si},
            state={Chungcheongbuk-do},
            country={Rep. of Korea}}

\author[2]{Yong-Goo Shin}[type=editor]
\cormark[1]

\ead{ygshin@hnu.kr}

\affiliation[2]{organization={Division of Smart Interdisciplinary Engineering, Hannam University},
            addressline={Daedeok-Gu}, 
            city={Daejeon},
            postcode={34430}, 
            country={Rep. of Korea}}

\cortext[1]{Corresponding author}


\begin{abstract}
This paper introduces a novel convolution method, called generative convolution (GConv), which is simple yet effective for improving the generative adversarial network (GAN) performance. Unlike the standard convolution, GConv first selects useful kernels compatible with the given latent vector, and then linearly combines the selected kernels to make latent-specific kernels. Using the latent-specific kernels, the proposed method produces the latent-specific features which encourage the generator to produce high-quality images. This approach is simple but surprisingly effective. First, the GAN performance is significantly improved with a little additional hardware cost. Second, GConv can be employed to the existing state-of-the-art generators without modifying the network architecture. To reveal the superiority of GConv, this paper provides extensive experiments using various standard datasets including CIFAR-10, CIFAR-100, LSUN-Church, CelebA, and tiny-ImageNet. Quantitative evaluations prove that GConv significantly boosts the performances of the unconditional and conditional GANs in terms of Inception score (IS) and Frechet inception distance (FID). For example, the proposed method improves both FID and IS scores on the tiny-ImageNet dataset from 35.13 to 29.76 and 20.23 to 22.64, respectively.
\end{abstract}



\maketitle

\section{Introduction}
Generative adversarial network (GAN)~\cite{goodfellow2014generative}, which allows to synthesize data distribution from a given prior distribution, has achieved great success for various applications such as text-to-image translation~\cite{reed2016generative, hong2018inferring}, image-to-image translation~\cite{isola2017image, choi2018stargan, zhu2017unpaired}, and image inpainting~\cite{yu2018free, sagong2019pepsi, shin2020pepsi++}. In the original setting, GAN consists of two different networks called generator and discriminator: the generator is trained to construct the real data distribution to fool the discriminator, whereas the discriminator aims to classify real samples from synthetic ones which are generated by the generator~\cite{wu2021gradient}. Since the generator and discriminator are trained to minimize opposing goals, \ie min-max game between them, training GAN stably is more difficult compared to supervised learning-based convolutional neural network (CNN)~\cite{zhang2019consistency}. Therefore, a recent line of study focuses on moderating the unstable training problem. Several works~\cite{zhang2019self, karras2017progressive, zhang2018stackgan++, brock2018large} presented novel generator and discriminator architectures. These methods generate high fidelity images that easily fool humans, but they cannot fully overcome the unstable training issue of GAN~\cite{park2021generative}. 

The sharp gradient space of the discriminator is the most popular reason for the unstable GAN training~\cite{wu2021gradient}. To mitigate this issue, some papers~\cite{miyato2018spectral, gulrajani2017improved, kodali2017convergence, zhang2019consistency} proposed regularization or normalization techniques that prevent the discriminator from making sharp gradients during the training procedure. As a regularization strategy, conventional methods~\cite{gulrajani2017improved, wu2019generalization, wei2018improving, kodali2017convergence, roth2017stabilizing} usually add regularization terms into the objective functions of adversarial learning. For instance, the gradient penalty-based regularization methods~\cite{gulrajani2017improved, wu2019generalization, wei2018improving}, which compute the gradient norm as the penalty term, have been widely used for regularization techniques. On the other hand, normalization-based methods~~\cite{miyato2018spectral, arjovsky2017wasserstein, kurach2019large} also have been widely studied. Spectral normalization (SN)~\cite{miyato2018spectral} is the most popular one that imposes the Lipschitz constraint on the discriminator around one by dividing weight matrices with the largest singular value. Since these regularization and normalization techniques successfully make the training procedure stably, most recent studies use these techniques in their applications. 

On the other hand, some researchers explored architectural modules to enhance the GAN performance. Miyato~\etal~\cite{miyato2018cgans} proposed a conditional projection module that employs the conditional information by projecting the conditional vector to the feature vectors. This approach significantly improves the quality of the class-conditional image generation. Zhang~\etal~\cite{zhang2019self} introduced a self-attention unit which guides the generator and discriminator where to attend. The self-attention unit can improve the GAN performance, but it requires high computational costs to produce the self-attention map due to the matrix multiplication operation. Yeo~\etal~\cite{yeo2021simple} introduced a simple yet effective technique, called a cascading rejection module, which produces dynamic features in an iterative manner at the last layer of the discriminator. Recently, Park~\etal~\cite{park2021GRB} proposed a novel residual block, called a generative residual block (GRB), which contains an additional side-residual path. By emphasizing the essential feature and suppressing the unnecessary one in the side-residual path, the GRB effectively decodes the latent feature and produces a high-quality image. 

However, there have been only a few attempts to redesign the standard convolution (Conv) to be effective for GAN training. Sagong~\etal~\cite{sagong2019cgans} proposed a novel conditional convolution (cConv) which incorporates the conditional information into the convolution operation. This approach effectively boosts the conditional GAN (cGAN) performance, but it has a limitation: cConv only can be used for cGAN scheme since it is designed for replacing the conditional batch normalization. Recently, Park~\etal~\cite{park2021generative} proposed perturbed convolution (PConv) that not only improves the GAN performance but also prevents the discriminator from falling into the overfitting problem. However, since PConv is developed for the discriminator, it is hard to apply to the generator. 

\begin{figure}
\centering
\includegraphics[width=0.9\linewidth]{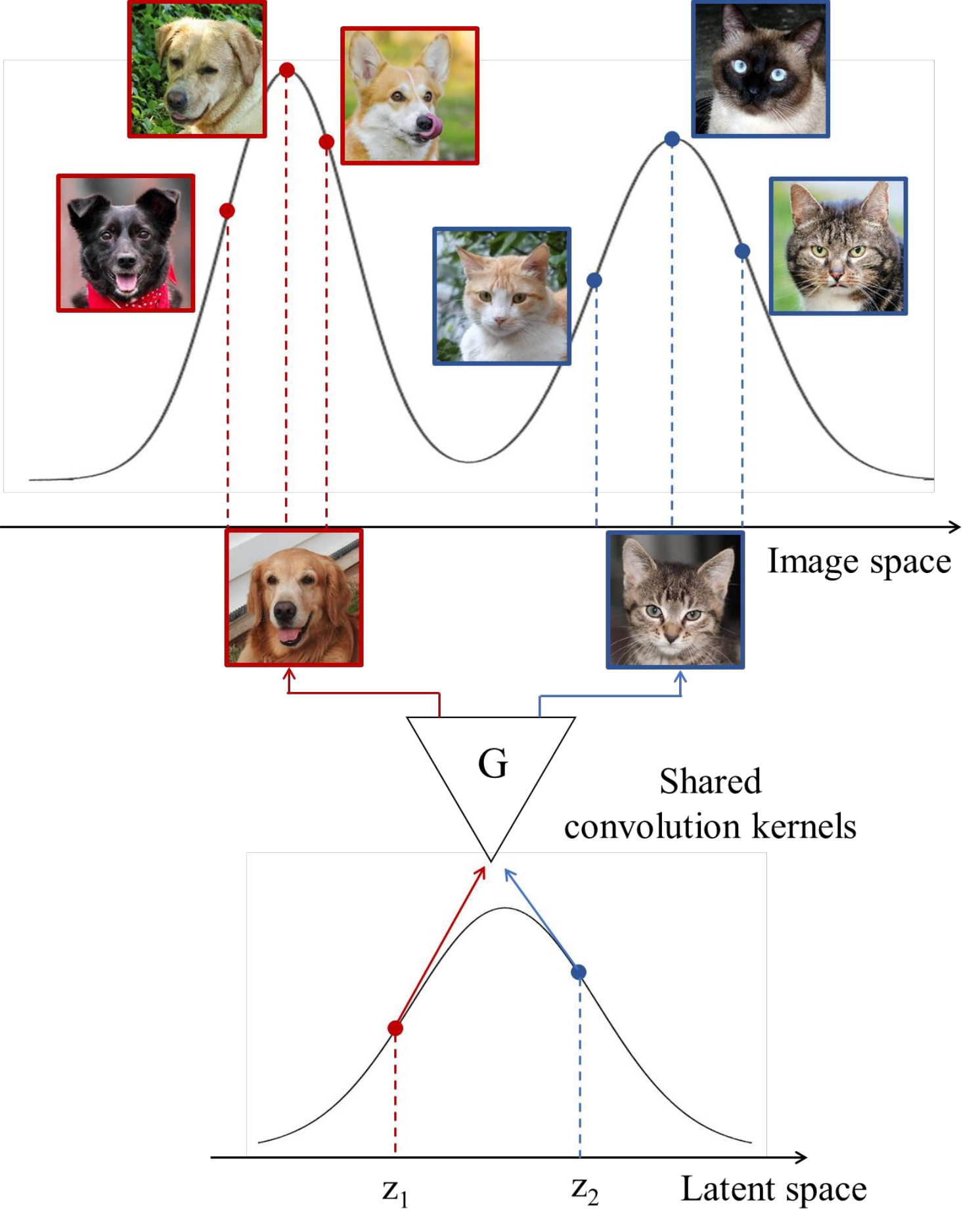}
\caption{The illustration of the major problem of the generator. The target image data is composed of two different modals, \ie cat and dog images. }
\label{fig:fig1}
\vspace{-0cm}
\end{figure}

Indeed, most state-of-the-art GAN studies~\cite{miyato2018spectral, miyato2018cgans, brock2018large, zhang2019self} construct their generator architectures by stacking multiple residual blocks~\cite{he2016deep, he2016identity} composed of Conv, batch normalization (BN)~\cite{ioffe2015batch}, and activation function. More specifically, these methods usually employed the latent vector such as the Gaussian random noise vector as the input of their generators. Thus, to produce different images according to the input latent vector, the generator should learn dynamic features that consider the input latent vector. However, since Conv uses the shared convolution kernel without considering the input latent vector, it is difficult to learn the dynamic features. We agree that even though Conv has this problem, it has shown fine performance in the GAN study. However, we believe that there would exist a novel convolution technique that is more suitable for the generator.

To prove our hypothesis, in this paper, we present a novel convolution method, called generative convolution (GConv), which effectively improves the GAN performance by learning the dynamic features considering the input latent vector, \ie latent-specific features. Different from Conv, GConv contains a simple kernel converting module that produces kernels representing the latent-specific features. More specifically, the kernel converting module first chooses useful kernels which are compatible with the input latent vector, and then linearly combines the selected kernels to produce latent-specific kernels. This approach is simple but surprisingly effective. First, with slight additional hardware costs, the GAN and cGAN performances are significantly improved. Second, since GConv has the same external structure with Conv, it can be easily implemented to the existing generators without altering the network architecture. To prove the superiority of GConv, we provide extensive experiments using various datasets such as CIFAR-10~\cite{krizhevsky2009learning}, CIFAR-100~\cite{krizhevsky2009learning}, LSUN-Church (LC)~\cite{yu15lsun}, CelebA (CA)~\cite{liu2015deep}, tiny-ImageNet (TIN)~\cite{23deng2009imagenet, yao2015tiny}, and CelebA-GQ~\cite{liu2015deep, karras2017progressive}. Quantitative evaluations show that GConv significantly boosts the GAN and cGAN performances in terms of Inception score (IS) and Frechet inception distance (FID). 

Key contributions of our paper are summarized as follows: First, we propose a novel convolution operation, \ie GConv, which is a direct replacement of the standard convolution. Second, extensive experiments on various datasets reveal that GAN and cGAN trained with GConv outperform those trained with Conv in terms of FID and IS. For instance, the proposed method improves FID and IS scores on the TIN dataset from 35.13 to 29.76 and 20.23 to 22.64, respectively.

\begin{figure}
\centering
\includegraphics[width=0.95\linewidth]{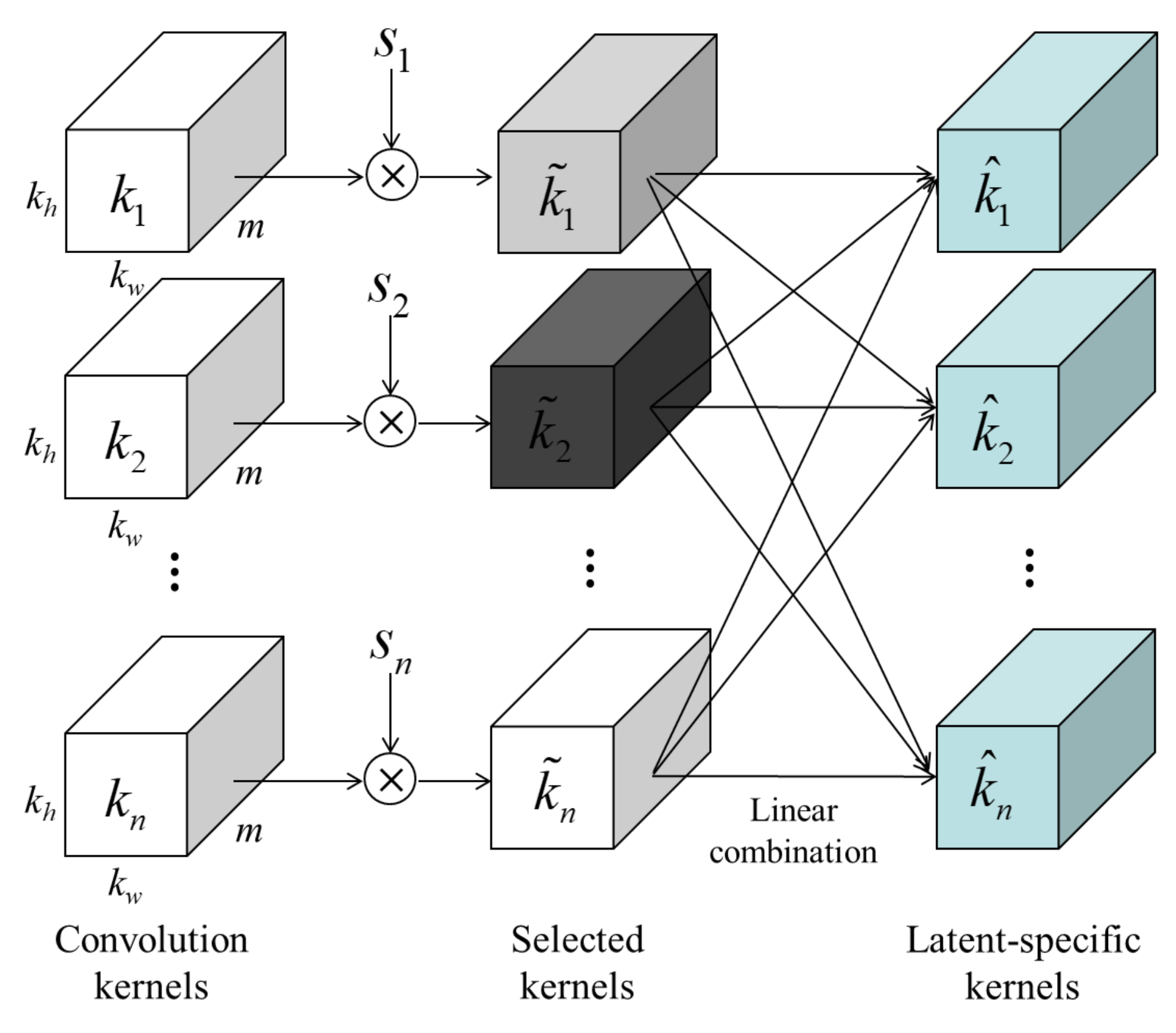}
\caption{The illustration of the kernel converting module.}
\label{fig:fig2}
\vspace{-0cm}
\end{figure}

\section{Preliminaries}
\label{sec2}
This section briefly introduces GAN and points out the limitations of existing methods. In general, the adversarial learning between the discriminator \textit{D} and the generator \textit{G} is formulated as 

\begin{eqnarray}
\label{eq1:ganD}
       \lefteqn{L_D = - E_{x\sim P_\textrm{data}(x)}[\log D(x)]}\nonumber\\
    & & {\qquad \qquad \qquad} - E_{z\sim P_{z(z)}}[\log(1-D(G(z)))],
\end{eqnarray}

\begin{eqnarray}
\label{eq1:ganG}
    L_G = -E_{z\sim P_{z(z)}}[\log(D(G(z)))],
\end{eqnarray}
where $L_D$ and $L_G$ are the objective functions for the $G$ and $D$, respectively. In addition, \textit{x} means a sample from the real data  $P_{data}(x)$ and \textit{z} indicates a latent vector sampled from latent distribution $P_z(z)$ such as Gaussian normal distribution. In this setting, $G$ is guaranteed to generate $P_{data}(x)$ if \textit{D} is always optimal. However, due to various issues such as gradient vanishing or explosion problems, this objective function is often fail to train both networks stably. To alleviate this problem, various studies have reformulated Eqs.~\ref{eq1:ganD} and~\ref{eq1:ganG}. Mao~\etal~\cite{mao2017least} proposed a least square error-based objective function (LSGAN), whereas Arjovsky~\etal~\cite{arjovsky2017wasserstein} employed Wasserstein distance between real and generated samples to build the objective function (WGAN). The current widely-used practice is training the GAN with the \textit{hinge} adversarial loss~\cite{yeo2021simple, miyato2018cgans, miyato2018spectral, brock2018large, park2021generative, sagong2019pepsi, shin2020pepsi++, chen2019self}. The \textit{hinge} adversarial loss is formulated as

\begin{figure*}
\centering
\includegraphics[width=0.85\linewidth]{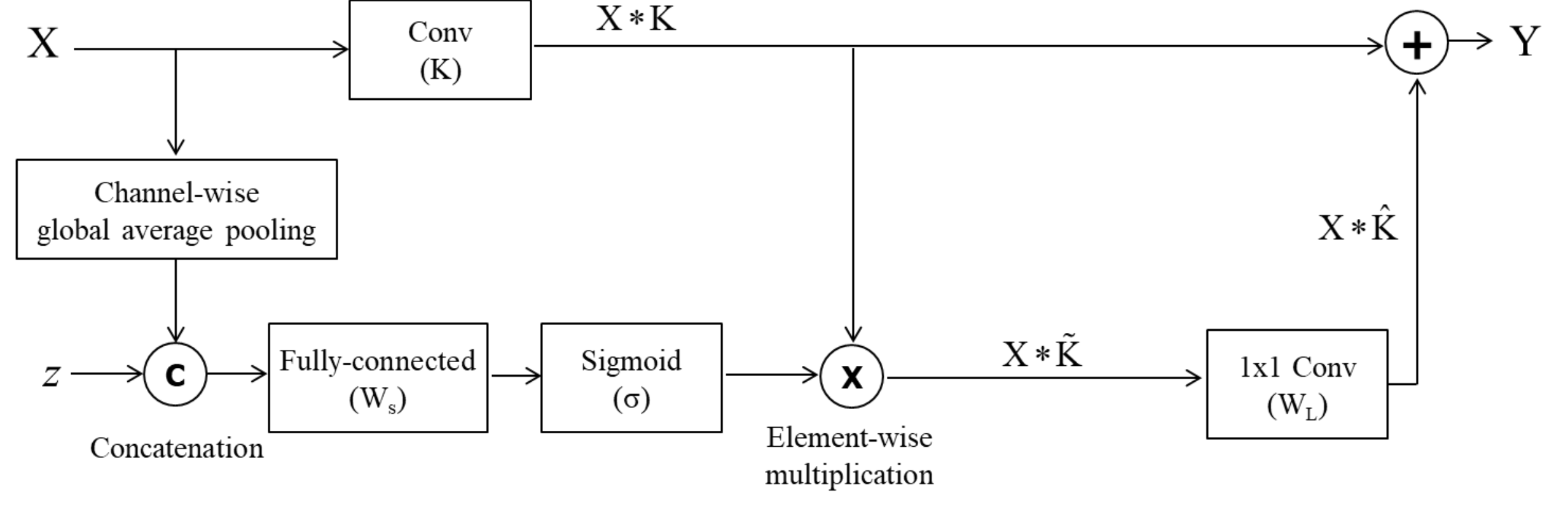}
\caption{The block diagram of GConv. Following the block diagram, GConv can be easily implemented by using deep-learning libraries such as \textit{Tensorflow} or \textit{Pytorch}.}
\label{fig:fig3}
\vspace{-0cm}
\end{figure*}

\begin{eqnarray}
\label{eq2_D}
    \lefteqn{L_D=E_{x\sim P_{data}(x)}[\max(0, 1-D(x))]} \nonumber\\
    & & { \qquad} + E_{z\sim P_{z(z)}}[max(0, 1+D(G(z)))],
\end{eqnarray}

\begin{equation}
    L_G=-E_{z\sim P_{z(z)}}[D(G(z))].
\label{eq2_G}
\end{equation}

To further improve the GAN performance as well as training stability, many studies have explored novel network architectures or normalization techniques. However, these methods still have a common issue with the generator. In detail, the generator is trained to synthesize the target data distribution, \ie $P_{data}(x)$. Thus, if $P_{data}(x)$ is a single-modal distribution, the generator makes samples easily following $P_{data}(x)$. However, if $P_{data}(x)$ consists of multiple modals, the generator is difficult to successfully generate all modals in $P_{data}(x)$. A major reason causing this problem is that the generator employs the shared convolution kernels without considering the input latent vector, \ie $z$. 

For ease of understanding, we prepare an example~\cite{park2021GRB}. Let us consider where $P_{data}(x)$ is a distribution that is composed of two classes of images, \ie dog and cat images, as shown in Fig.~\ref{fig:fig1}. In our example, we assume that the left and right sides of the latent space are mapped to the dog and cat images, respectively. To produce the visually plausible images, the generator should learn features specialized for each class. However, since the generator used the shared convolutional kernels, it is difficult to produce dynamic features specialized for each class. That means it is needed to learn features that consider $z$ for generating the diverse images well. To alleviate this problem, this paper proposes a novel convolution technique which is effective to GAN. 

\section{Proposed Method}
\label{sec3}
\subsection{Generative Convolution (GConv)}
\label{subsec3.1}
This paper presents a novel convolution operation, \ie GConv, which produces latent-specific features. In GConv, there is a simple kernel converting module that modulates convolution kernels for learning the latent-specific features. Fig.~\ref{fig:fig2} illustrates the kernel converting module. As shown in Fig.~\ref{fig:fig2}, the input of the kernel converting module is the general convolution kernel $\textrm{K}=\{k_1, ..., k_n\}\in \mathbb{R}^{o \times n}$, where \textit{o} indicates the number of parameters in each kernel, \ie $o = k_h \times k_w \times m$ in which $k_h$, $k_w$ and $m$ indicate the kernel height, width, and input feature dimension, respectively, and \textit{n} means the number of kernels. In the kernel converting module, the network first selects kernels by multiplying the scaling value $s_i$ which is computed as follows:
\begin{equation}
    \textrm{S}=\sigma((x_m\odot z)^\textrm{T}\textrm{W}_\textrm{s}),
\label{eq4}
\end{equation}
where $z\in \mathbb{R}^{1 \times d_z}$ is the input latent vector with $d_z$ dimension and $x_m\in \mathbb{R}^{1 \times m}$ is a mean vector obtained by applying channel-wise global average pooling in the input tensor. $W_s\in \mathbb{R}^{(m + d_z)\times n}$ is a weight matrix trained for inferring the scaling parameters $\textrm{S}=\{s_1, ..., s_n\}\in \mathbb{R}^{1 \times n}$. In addition, $\sigma$ and $\odot$ indicate a sigmoid function and concatenation operation, respectively. Using Eq.~\ref{eq4}, each GConv individually computes the scaling parameters in the range zero to one. By multiplying $s_i$ to each kernel $k_i$, the generator pays less attention to the useless kernels needed for producing images. In other words, by using \textrm{S}, the generator could build a set of selected kernels $\tilde{\textrm{K}} \in \mathbb{R}^{o \times n}$ that considers the latent vector. 

After building $\tilde{\textrm{K}}$, the latent-specific kernels $\hat{\textrm{K}}\in \mathbb{R}^{o \times n}$ can be produced by linearly-combining the $\tilde{\textrm{K}}$, which is defined as
\begin{equation}
    \hat{\textrm{K}}=\tilde{\textrm{K}}\textrm{W}_\textrm{L},
\label{eq5}
\end{equation}
where $\textrm{W}_\textrm{L}\in \mathbb{R}^{n \times n}$ is the weight matrix trained for producing $\hat{\textrm{K}}$. It is worth noting that all operations in Eqs.~\ref{eq4} and~\ref{eq5} are simple linear algebraic computations; the proposed method requires a small number of additional computational costs, \ie $\textrm{W}_\textrm{s}$ and $\textrm{W}_\textrm{L}$. Using the latent-specific kernels, the output features of GConv is naturally computed as follows:

\begin{equation}
    \textrm{Y}=\textrm{X}\ast \textrm{K} + \textrm{X}\ast \hat{\textrm{K}},
\label{eq6}
\end{equation}
where $\textrm{X}\in\mathbb{R}^{b\times h\times w\times m}$ and $\textrm{Y}\in\mathbb{R}^{b\times h\times w\times n}$ are input and output features, and \textit{b}, \textit{h}, and \textit{w} indicate the sizes of batch, height, and width, respectively. Here, Eq.~\ref{eq6} can be reformulated by replacing the $\hat{K}$ term as follows:

\begin{equation} 
\label{eq7}
\begin{split}
\textrm{Y} & =\textrm{X}\ast \textrm{K} + \textrm{X}\ast\hat{\textrm{K}}\\
 & = \textrm{X}\ast \textrm{K} + \textrm{X}\ast \tilde{\textrm{K}}W_L\\
 & = \textrm{X}\ast \textrm{K} + \textrm{X}\ast \textrm{K}\textrm{Diag(\textrm{S})}\textrm{W}_\textrm{L}\\
\end{split}
\end{equation}
where Diag($\cdot$) is a function creating a diagonal matrix from the given vector. It is worth noting that the second term in Eq.~\ref{eq7}, \ie $\textrm{X}\ast \textrm{K}\textrm{Diag(\textrm{S})}\textrm{W}_\textrm{L}$, can be easily represented with the neural network operation. More specifically, the Diag(S) multiplication acts the same role with element-wise multiplication, whereas the $\textrm{W}_\textrm{L}$ multiplication is equivalent to $1\times1$ convolution operation. Based on these observations, we draw a block diagram of GConv as depicted in Fig.3. This block diagram has the mathematically same meaning with Eq.~\ref{eq7}. Therefore, it is easy to implement GConv by using deep-learning libraries such as \textit{Tensorflow} or \textit{Pytorch}.

\begin{figure*}
\centering
\includegraphics[width=0.95\linewidth]{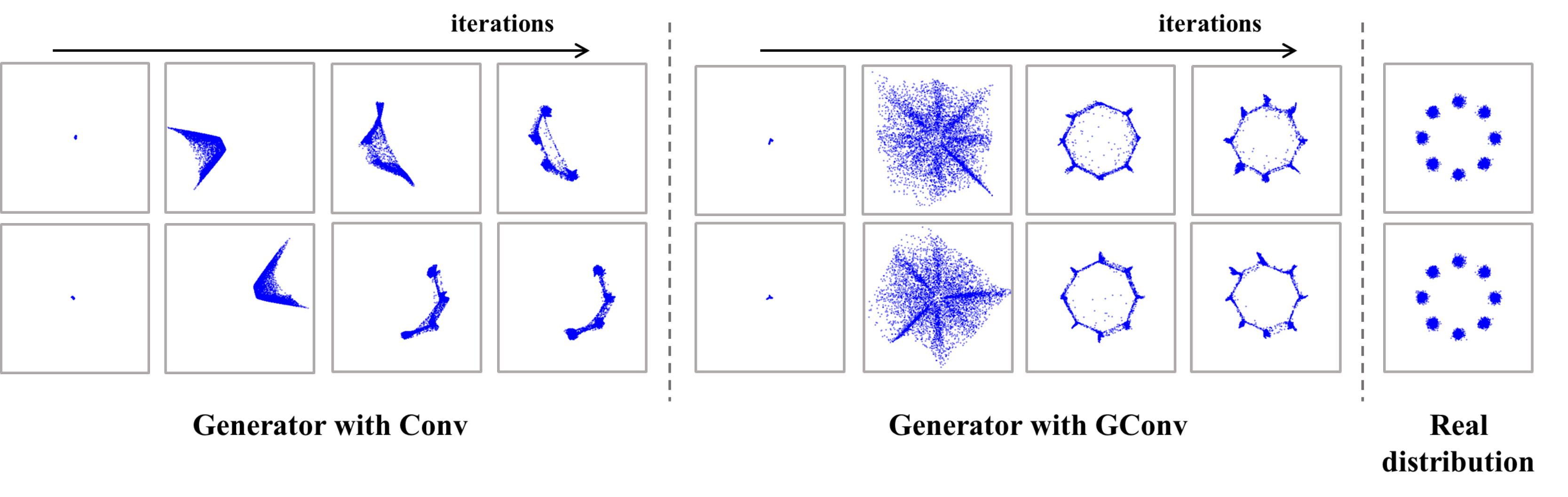}
\caption{Illustration of the GAN training results on eight 2D Gaussian mixture models. We trained the networks two times to reveal the effectiveness of GConv.}
\label{fig:fig_toy}
\vspace{-0cm}
\end{figure*}

To clarify the effectiveness of the proposed method, we trained a GAN with a simple network architecture consisting of multiple fully-connected layers. In our experiments, eight two-dimensional (2D) Gaussian mixture models (GMMs) are used as the real data distribution. For a fair comparison, in the generator, we only replaced the standard fully-connected layers with the GConv. Fig.~\ref{fig:fig_toy} illustrates the experimental results. The generator with standard fully-connected layers suffers from the mode collapse problem, whereas the generator with the GConv generates all the GMMs successfully. These results demonstrate that the proposed method is effective to train GAN. More extensive experiments will be presented in Sec.~\ref{subsec:4.2}. 

Indeed, one may anticipate that GConv is similar to the conventional channel-attention (CA) methods~\cite{woo2018cbam, wang2020eca}, which emphasizing the selected channels of the input feature. However, there is a major difference between the proposed method and conventional ones: the proposed method is an alternative operation of convolution, but the CA technique is a post-processing algorithm used after conducting the convolution operation. More specifically, existing CA methods are usually located after performing the convolution operation to enhance the specific channels of output feature. In contrast, the proposed method directly selects and combines kernels in the convolution layer. That means it is able to use the existing CA method after performing GConv. The detailed explanations will be described in the section~\ref{subsec:4.2}. 

\begin{figure}
\centering
\includegraphics[width=0.95\linewidth]{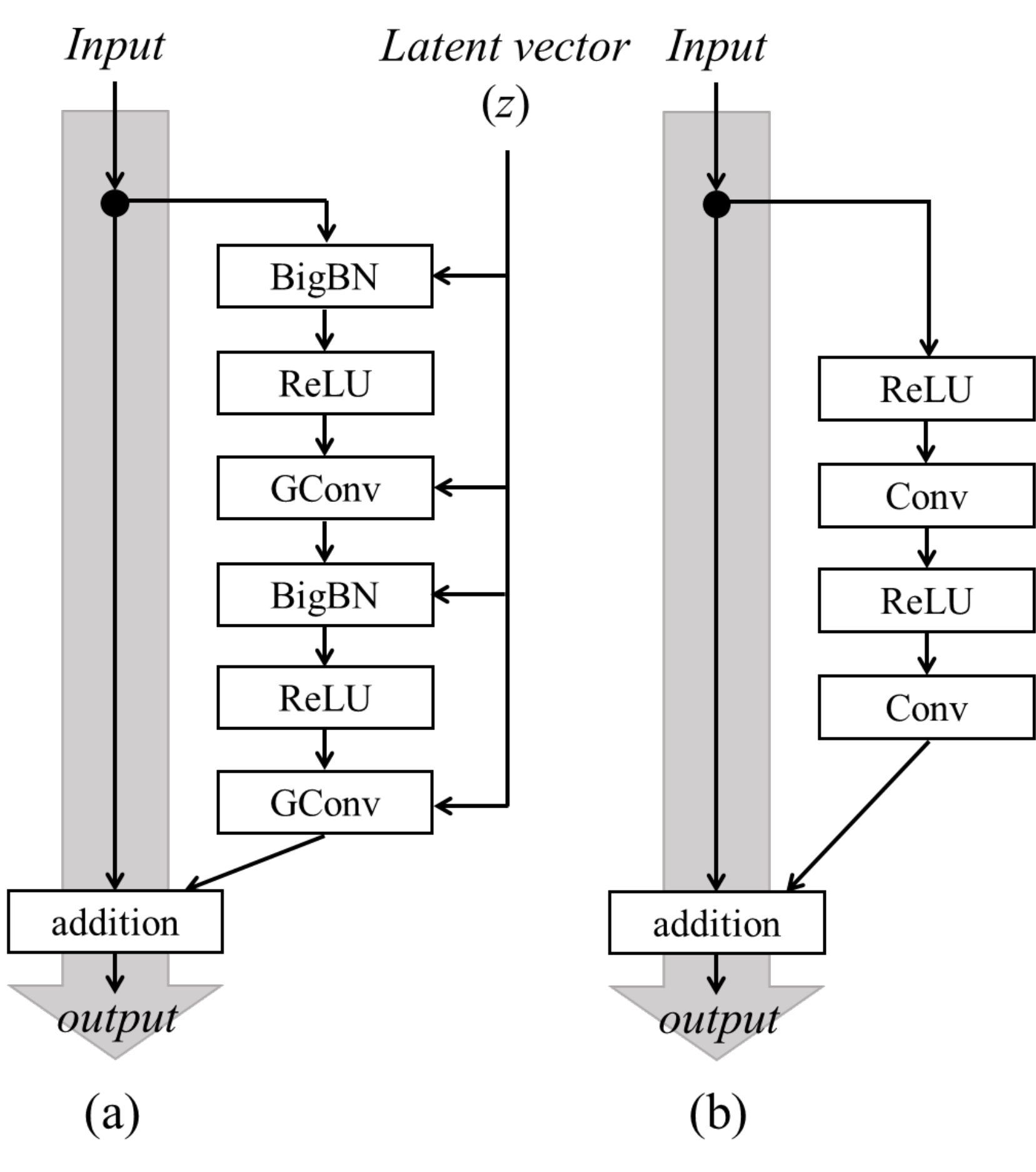}
\caption{The illustration of the RBs in the generator and the discriminator. (a) RB in the generator, (b) RB in the discriminator.}
\label{fig:fig4}
\vspace{-0cm}
\end{figure}

\subsection{Implementation Details}
\label{subsec:3.2}
To prove the superiority of GConv, we conducted extensive experiments utilizing the various datasets which are widely used for evaluating the GAN performance: CIFAR-10~\cite{krizhevsky2009learning}, CIFAR-100~\cite{krizhevsky2009learning}, LC~\cite{yu15lsun}, CA~\cite{liu2015deep}, TIN~\cite{23deng2009imagenet, yao2015tiny}, and CelebA-HQ~\cite{liu2015deep, karras2017progressive}. The resolutions of the CIFAR-10 and CIFAR-100 datasets are $32\times 32$, whereas the images of LC, CA, and TIN datasets are resized to $128\times 128$ pixels. Also, the CelebA-HQ dataset is used for training the networks that produce the $256\times 256$ images. To train GAN, we used the hinge-version loss in Eqs.~\ref{eq2_D} and~\ref{eq2_G} as the objective function. Since all parameters in the discriminator and generator including GConv can be differentiated, we employed the Adam optimizer~\cite{kingma2014adam} and set the user parameters of Adam optimizer, \ie $\beta _1$ and $\beta _2$, to 0 and 0.9, respectively.

For training the CIFAR-10, CIFAR-100, and CelebA-HQ datasets, we set the learning rate as 0.0002, and the discriminator was updated 5 times using different mini-batches when the generator is updated once. In contrast, for training the LC, CA, and TIN, we employed a two-time scale update rule (TTUR)~\cite{heusel2017gans} that the learning rates of the generator and discriminator are set to 0.0001 and 0.0004, respectively. Note that the TTUR technique updates the discriminator a single time when the generator is updated once. For all datasets, we decreased the learning rate linearly in the last 50,000 iterations. For the CIFAR-10 and CIFAR-100 datasets, we set a batch size of the discriminator as 64 and trained the generator for 50k iterations. In our experiments, following the previous papers~\cite{miyato2018cgans, miyato2018spectral, park2021GRB}, the generator was trained with a batch size twice as large as when training the discriminator. That means the generator and discriminator were trained with 128 and 64 batch size, respectively. For the LC, CA, and TIN datasets, both batch sizes of the discriminator and the generator are set to 32. The network is trained to 300k iterations on LC and CA datasets and 1M iterations on TIN dataset. For the CelebA-HQ dataset, the network is trained to 100k iterations, while setting the batch sizes of the generator and discriminator to 32 and 16, respectively.

\begin{table}[t]
\caption{Network architecture of the generator for each image resolution. The input latent vector is sample from $z \in \mathbb{R}^{32} \sim N(0, I)$}
\begin{center}
\begin{tabular}{c | c | c }
\hline\hline
$32\times32$ image& $128\times128$ image& $256\times256$ image \\
\hline
FC, $4 \times 4 \times 256$ & FC, $4 \times 4 \times 512$ & FC, $4 \times 4 \times 512$ \\
RB, up, 256 & RB, up, 512 & RB, up, 512 \\
RB, up, 256 & RB, up, 512 & RB, up, 512 \\
RB, up, 256 & RB, up, 256 & RB, up, 256 \\
BN, ReLU & RB, up, 128 & RB, up, 128 \\
$3\times3$ conv, Tanh & RB, up, 64 & RB, up, 64 \\
& BN, ReLU & RB, up, 32 \\
& $3\times3$ conv, Tanh & BN, ReLU  \\
& & $3\times3$ conv, Tanh \\

\hline\hline
\end{tabular}
\end{center}
\label{table:table_G}
\end{table}

\begin{table}[t]
\caption{Network architecture of the discriminator for each image resolution.}
\begin{center}
\begin{tabular}{c | c | c }
\hline\hline
$32\times32$ image & $128\times128$ image & $256\times256$ image\\
\hline
RB, down, 128 & RB, down, 64 & RB, down, 32 \\
RB, down, 128 & RB, down, 128 & RB, down, 64 \\
RB, 128 & RB, down, 256 & RB, down, 128 \\
RB, 128 & RB, down, 512 & RB, down, 256 \\
ReLU & RB, down, 512 & RB, down, 512 \\
Global Sum & RB, 512 & RB, down, 512 \\
Dense, 1 & ReLU & RB, 512 \\
 & Global Sum & ReLU \\
 & Dense, 1 & Global Sum \\
 & & Dense, 1 \\

\hline\hline
\end{tabular}
\end{center}
\label{table:table_D}
\end{table}

\begin{table*}[t]
\caption{Comparison of GConv with the standard convolution on CIFAR-10 and CIFAR-100 datasets in terms of FID and IS.}
\begin{center}
\begin{tabular}{c | c | c | c | c | c | c  | c | c | c | c | c | c}
\hline
& \multirow{2}*{Dataset} & & \multicolumn{2}{c|}{Conv} & \multicolumn{2}{c|}{Conv*} & \multicolumn{2}{c|}{$\textrm{Conv}^{\dagger}$} & \multicolumn{2}{c|}{GConv} & \multicolumn{2}{c}{ $\textrm{GConv}^{\dagger}$}\\
\cline{4-13}
& & & IS$\uparrow$ & FID$\downarrow$ & IS$\uparrow$ & FID$\downarrow$ & IS$\uparrow$ & FID$\downarrow$ & IS$\uparrow$ & FID$\downarrow$ & IS$\uparrow$ & FID$\downarrow$ \\
\hline

\multirow{8}*{GAN} & \multirow{4}*{CIFAR-10} & trial 1 & 7.70 & 13.98 & 7.81 & 13.22 & 8.14 & 11.70 & 7.80 & 12.93 & 8.17 & 11.69\\
& & trial 2 & 7.80 & 12.69 & 7.75 & 12.80 & 8.04 & 12.78 & 7.95 & 12.51 & 8.26 & 12.23\\
& & trial 3 & 7.87 & 13.09 & 7.69 & 13.77 & 8.14 & 11.97 & 7.93 & 12.62 & 8.27 & 11.48\\
\cline{3-13}
& & \textbf{Average} & \textbf{7.79}  & \textbf{13.58} & \textbf{7.75} & \textbf{13.26} & \textbf{8.11} & \textbf{12.15} & \textbf{7.89} & \textbf{12.69} & \textbf{8.23} & \textbf{11.80}\\
\cline{2-13}

& \multirow{4}*{CIFAR-100} & trial 1 & 7.99 & 17.45 & 7.93 & 17.68 & 8.29 & 16.02 & 8.25 & 15.89 & 8.32 & 16.26\\
&  & trial 2 & 7.94 & 17.72 & 8.12 & 16.77 & 8.24 & 16.58 & 8.13 & 16.28 & 8.31 & 15.56\\
& & trial 3 & 7.94 & 17.57 & 8.01 & 17.36 & 8.28 & 16.00 & 8.14 & 16.47 & 8.26 & 15.28\\
\cline{3-13}
&  & \textbf{Average} & \textbf{7.96}  & \textbf{17.58} & \textbf{8.02} & \textbf{17.27} & \textbf{8.27} & \textbf{16.20} & \textbf{8.17} & \textbf{16.21} & \textbf{8.30} & \textbf{15.70}\\
\hline
\hline

\multirow{8}*{cGAN} & \multirow{4}*{CIFAR-10} & trial 1 & 8.09 & 9.36 & 8.10 & 9.61 & 8.41 & 8.67 & 8.25 & 8.73 & 8.52 & 8.03 \\
& & trial 2 & 7.97 & 9.62 & 8.16 & 9.46 & 8.32 & 8.90 & 8.31 & 9.03 & 8.59 & 8.17\\
& & trial 3 & 8.02 & 9.38 & 8.01 & 9.61 & 8.39 & 8.22 & 8.33 & 8.58 & 8.48 & 8.05\\
\cline{3-13}
& & \textbf{Average} & \textbf{8.03}  & \textbf{9.45} & \textbf{8.09} & \textbf{9.56} & \textbf{8.37} & \textbf{8.60} & \textbf{8.30} & \textbf{8.78} & \textbf{8.53} & \textbf{8.08}\\
\cline{2-13}

& \multirow{4}*{CIFAR-100} & trial 1 & 8.88 & 13.17 & 8.90 & 12.41 & 9.40 & 11.49 & 9.20 & 11.48 & 9.63 & 10.50 \\
&  & trial 2 & 8.89 & 13.20 & 9.02 & 12.80 & 9.35 & 11.81 & 9.18 & 11.53 & 9.57 & 10.16\\
& & trial 3 & 8.87 & 13.01 & 8.85 & 12.91 & 9.45 & 11.76 & 9.13 & 11.77 & 9.53 & 11.11\\
\cline{3-13}
&  & \textbf{Average} & \textbf{8.88}  & \textbf{13.13} & \textbf{8.93} & \textbf{12.71} & \textbf{9.40} & \textbf{11.69} & \textbf{9.17} & \textbf{11.60} & \textbf{9.58} & \textbf{10.59}\\
\hline

\end{tabular}
\end{center}
\label{table:table2}
\end{table*}

To quantify the importance of GConv, we employed a strong baseline in~\cite{miyato2018spectral} following the state-of-the-art studies~\cite{miyato2018cgans, brock2018large, zhang2019self, park2021GRB}. More specifically, we employed the generator and discriminator architectures constructing with multiple residual blocks (RBs) as our baseline models; we just replaced Conv in the generator with GConv and did not alter the architecture of the discriminator. Note that we used the improved BN introduced in BigGAN~\cite{brock2018large} which infers affine transformation parameters from \textit{z}. In this paper, we call the improved BN as BigBN. Therefore, in each residual block, \textit{z} is employed to BigBN and GConv. The detailed architecture of RB is illustrated in Fig.~\ref{fig:fig4}. 

For training the CIFAR-10 and CIFAR-100 datasets the spectral normalization (SN)~\cite{miyato2018spectral} is only used for the discriminator, whereas the SN is applied to both generator and discriminator for training LC, CA, and TIN datasets~\cite{zhang2019self}. In the discriminator, the feature maps are down-sampled by utilizing the average-pooling after the second convolution. In the generator, the up-sampling (a nearest-neighbor interpolation) operation is located before the first convolution. A detailed descriptions of the network architectures for the generator and discriminator are described in Tables~\ref{table:table_G} and~\ref{table:table_D}. To train the network in the conditional GAN (cGAN) framework, following the most representative cGAN scheme, we replaced the BigBN in the generator with the conditional BigBN~\cite{brock2018large} and added the conditional projection layer in the discriminator~\cite{miyato2018cgans}. More specifically, in the RB of the generator, we concatenated \textit{z} with the one-hot vector which represents the input class, and used the concatenated vector to conditional BigBN and GConv. Note that other network architectures were the same as the models used for experiments of unconditional GAN.

\section{Experiments}
\label{sec4}
\subsection{Evaluation Metric}
\label{subsec:4.1}
In the field of GAN, FID~\cite{heusel2017gans} and IS~\cite{salimans2016improved} are most widely used assessments that evaluate how well the generator produces image. More specifically, FID is defined as follows: 

\begin{equation}
    \textrm{F}(p,q) = \| \mu_p - \mu_q \|_2^2 + \mathrm{trace}(C_p +C_q - 2(C_p C_q)^{\frac{1}{2}}),
\end{equation}
where $ \{\mu_p,C_p \}$ and $\{\mu_q,C_q \}$ are the mean and covariance of the samples with distributions of real and generated images, respectively. Lower FID scores mean better quality of the generated images. To measure the performance using FID, in this paper, we generated 50,000 images for each dataset. On the other hand, IS is expressed as 

\begin{equation}
I = exp(E[D_{KL}(p(l|X)||p(l))]),
\end{equation}
where \textit{l} is the label predicted by the Inception model~\cite{26szegedy2016rethinking} trained for classifying ImageNet dataset~\cite{23deng2009imagenet} having a thousand classes, and $p(l|X)$ and $p(l)$ represent the conditional class distributions and marginal class distributions, respectively. That means it measures how the generator produces diverse images with high-quality. However, IS has an one major drawback: IS is no meaning when evaluating the performance in a single class dataset such as CA, since it measures the score using the KL-divergence between $p(l|X)$ and $p(l)$. Therefore, in this paper, we did not use IS for evaluating the GAN performance for CA and LSUN datasets.

\begin{figure}
\centering
\includegraphics[width=0.95\linewidth]{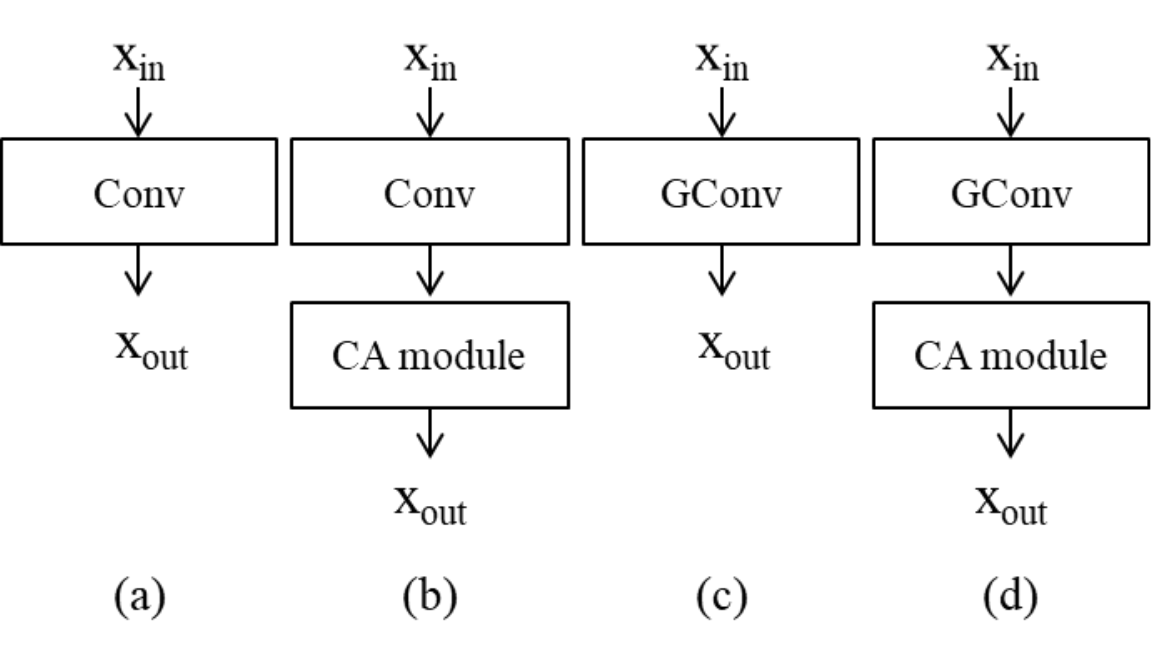}
\caption{The illustrations of Conv and GConv variations in Table~\ref{table:table2}. (a) Conv, (b) $\textrm{Conv}^{\dagger}$, (c) GConv, (d) $\textrm{GConv}^{\dagger}$.}
\label{fig:fig_table}
\vspace{-0cm}
\end{figure}

\subsection{Experimental Results}
\label{subsec:4.2}
Before proving the superiority of GConv on various datasets, we first conducted extensive ablation studies on CIFAR-10 and CIFAR-100 datasets. In our experiments, the network was trained three times from scratch to prove that the performance improvement was not due to the lucky weight initialization. First, we compared the GAN performances between Conv and GConv. As described in Table~\ref{table:table2}, the proposed method exhibits better performance than Conv in terms of FID and IS. It is worth noting that the proposed method shows superior performance to Conv in the cGAN as well as GAN. That means the GConv could be used for both studies for improving the performance. In addition, the proposed method consistently outperforms its counterpart, in both CIFAR-10 and CIFAR-100 datasets. These observations indicate that the latent-specific features produced in GConv effectively support the generator to produce high-quality images. 

\begin{table}[t]
\caption{Comparison of IS and FID scores when training the networks with various loss functions on the CIFAR-10 dataset.}
\begin{center}
\begin{tabular}{c | c | c | c | c | c}
\hline
\multirow{2}*{Loss function} & & \multicolumn{2}{c|}{Conv} & \multicolumn{2}{c|}{GConv} \\
\cline{2-6}
 & & IS$\uparrow$ & FID$\downarrow$ & IS$\uparrow$ & FID$\downarrow$ \\
\hline
\multirow{4}*{CE~\cite{goodfellow2014generative}} & trial 1 & 7.53 & 17.39 & 7.74 & 16.44 \\
 & trial 2 & 7.50 & 18.06 & 7.75 & 16.49 \\
 & trial 3 & 7.51 & 16.78 & 7.65 & 16.85 \\
\cline{2-6}
 & \textbf{Average} & 7.53 & 17.39 & \textbf{7.71} & \textbf{16.60} \\
 
\hline
\multirow{4}*{LSGAN~\cite{mao2017least}} & trial 1 & 7.20 & 21.30 & 7.44 & 18.05 \\
 & trial 2 & 7.15 & 19.67 & 7.47 & 17.65 \\
 & trial 3 & 7.16 & 20.73 & 7.53 & 17.65 \\
\cline{2-6}
 & \textbf{Average} & 7.17 & 20.57 & \textbf{7.48} & \textbf{17.78} \\

\hline
\end{tabular}
\end{center}
\label{table:table_loss}
\end{table}

Indeed, owing to $\textrm{W}_\textrm{s}$ and $\textrm{W}_\textrm{L}$ in Eqs.~\ref{eq4} and~\ref{eq5}, GConv needs a slightly more network parameters than Conv. For instance, when training the CIFAR-10 and CIFAR-100 datasets, the generator with Conv uses about 3.54M convolution weights, whereas that with GConv utilizes about 4.37M convolution weights. Thus, one may anticipate that the performance improvement is caused due to the increased number of network parameters. To mitigate this issue, we conducted ablation studies that match the computational costs of Conv with that of GConv. More specifically, we increased the number of output channels in Conv to match the number for convolution weights with GConv. In this paper, we referred the Conv with increased output channels as Conv*. As described in Table~\ref{table:table2}, Conv* shows performance either slightly improved or failed to improve even increasing the number of network parameters. In particular,  on the CIFAR-10 dataset, Conv* trained with the cGAN framework shows a poor FID score compared with Conv. In addition, GConv still shows better performance than Conv*. These results reveal that even if the number of network parameters is increased, it is difficult to further improve the performance using Conv. In contrast, GConv significantly boosts the GAN performance successfully in both GAN and cGAN. Based on these observations, we concluded that GConv has a fine trade-off between improved performance and increased computational costs, which makes GConv to be fine for practical use. 

On the other hand, one may have a query that the difference between GConv and conventional CA modules. As mentioned in Section~\ref{subsec3.1}, the major difference between the two methods is their locations in the network: GConv is a replacement of Conv, whereas the CA module is the post-processing method used after performing convolution operation. In other words, the existing CA techniques can be used after GConv. To clarify this claim, we conducted ablation studies that attach the CA module after Conv and GConv. In our experiments, we used the CA technique introduced in~\cite{woo2018cbam}, which is the most popular one among the CA techniques. In this paper, we denoted Conv and GConv with CA module as $\textrm{Conv}^{\dagger}$ and $\textrm{GConv}^{\dagger}$, respectively. The illustrations about $\textrm{Conv}^{\dagger}$ and $\textrm{GConv}^{\dagger}$ are depicted in Fig.~\ref{fig:fig_table}. As described in Table~\ref{table:table2} $\textrm{Conv}^{\dagger}$ shows better performance than Conv, which means the exiting CA module is effective to improve the GAN performance. It is worth noting that $\textrm{GConv}^{\dagger}$ also shows better performance than GConv. In addition, $\textrm{GConv}^{\dagger}$ exhibits superior performance to $\textrm{Conv}^{\dagger}$ in both GAN and cGAN schemes. These results contain two important meanings. First, GConv can be generally used as the replacement of Conv since it is compatible with the existing techniques designed for Conv. Second, GConv and the CA module act different roles in the networks; they can be used together for improving the GAN performance. Based on these observations, we concluded that it does not need to compare the effectiveness between GConv and the CA module. Therefore, in the remainder of this paper, we only compared the performance of GConv with that of Conv.

\begin{table}[t]
\caption{Comparison of IS and FID scores when training the networks with various activation functions on the CIFAR-10 dataset.}
\begin{center}
\begin{tabular}{c | c | c | c | c | c}
\hline
Activation & & \multicolumn{2}{c|}{Conv} & \multicolumn{2}{c|}{GConv} \\
\cline{2-6}
function & & IS$\uparrow$ & FID$\downarrow$ & IS$\uparrow$ & FID$\downarrow$ \\
\hline
\multirow{4}*{ELU~\cite{clevert2015fast}} & trial 1 & 7.69 & 14.87 & 8.19 & 11.83 \\
 & trial 2 & 7.60 & 14.99 & 7.97 & 12.25 \\
 & trial 3 & 7.68 & 15.38 & 8.17 & 12.80 \\
\cline{2-6}
 & \textbf{Average} & 7.65 & 15.08 & \textbf{8.11} & \textbf{12.29} \\
 
\hline
\multirow{4}*{GLU~\cite{dauphin2017language}} & trial 1 & 7.96 & 11.05 & 8.12 & 10.18 \\
 & trial 2 & 7.93 & 11.52 & 8.11 & 10.13 \\
 & trial 3 & 7.92 & 11.10 & 8.28 & 10.25 \\
\cline{2-6}
 & \textbf{Average} & 7.94 & 11.23 & \textbf{8.17} & \textbf{10.19} \\

\hline
\end{tabular}
\end{center}
\label{table:table_AC}
\end{table}

Furthermore, to show the generalization ability of GConv, we conducted other ablation studies in which the networks are trained with different experimental settings. First, we trained the networks using various adversarial objective functions: the objective function based on the cross-entropy (CE) theorem (Eqs.~\ref{eq1:ganD} and~\ref{eq1:ganG}) and the objective function proposed in the least square GAN (LSGAN) paper~\cite{mao2017least}. As shown in Table~\ref{table:table_loss}, the proposed method outperforms Conv even trained with various objective functions. That means GConv can be generally applied to train the GAN regardless of the objective function. In addition, to check the compatibility of GConv with various activation functions, we changed the activation function in the RBs of the generator. In our experiments, we employed the exponential linear unit (ELU)~\cite{clevert2015fast} and gated linear unit (GLU)~\cite{dauphin2017language} as the activation function. As summarized in table~\ref{table:table_AC}, the proposed method exhibits superior performance to Conv in the various activation functions. Based on these results, we expected that the GConv can be generally used for the GAN-based image generation techniques having various experimental settings. 

\begin{figure*}
\centering
\includegraphics[width=0.85\linewidth]{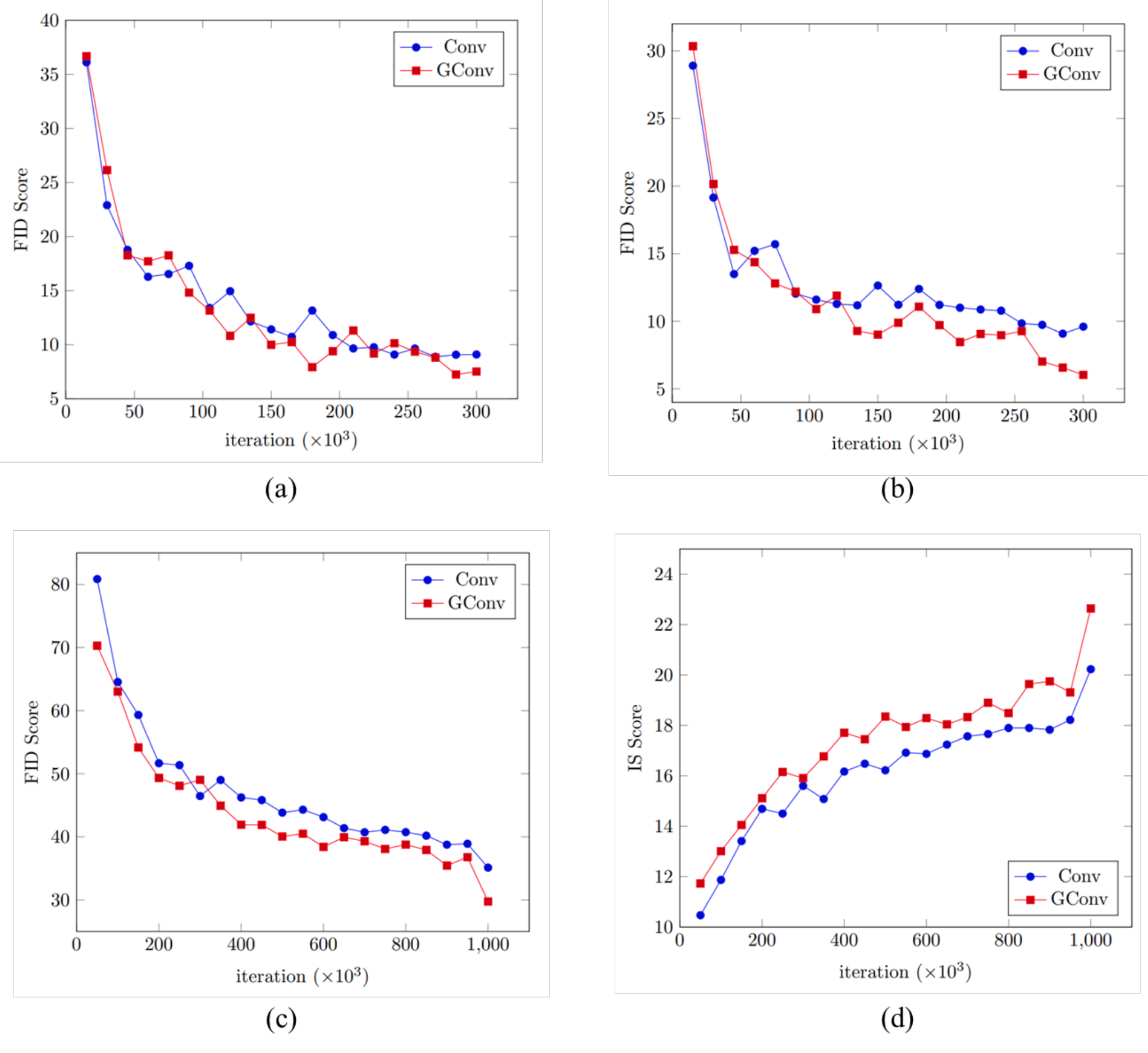}
\caption{The FID and IS score graphs. The blue line indicates GConv, whereas the red one means Conv. (a) FID score graph on the LC dataset, (b) FID score graph on the CA dataset, (c) FID score graph on the TIN dataset, (d) IS score graph on the TIN dataset.}
\label{fig:fig5}
\vspace{-0cm}
\end{figure*}

\begin{figure*}
\centering
\includegraphics[width=1.0\linewidth]{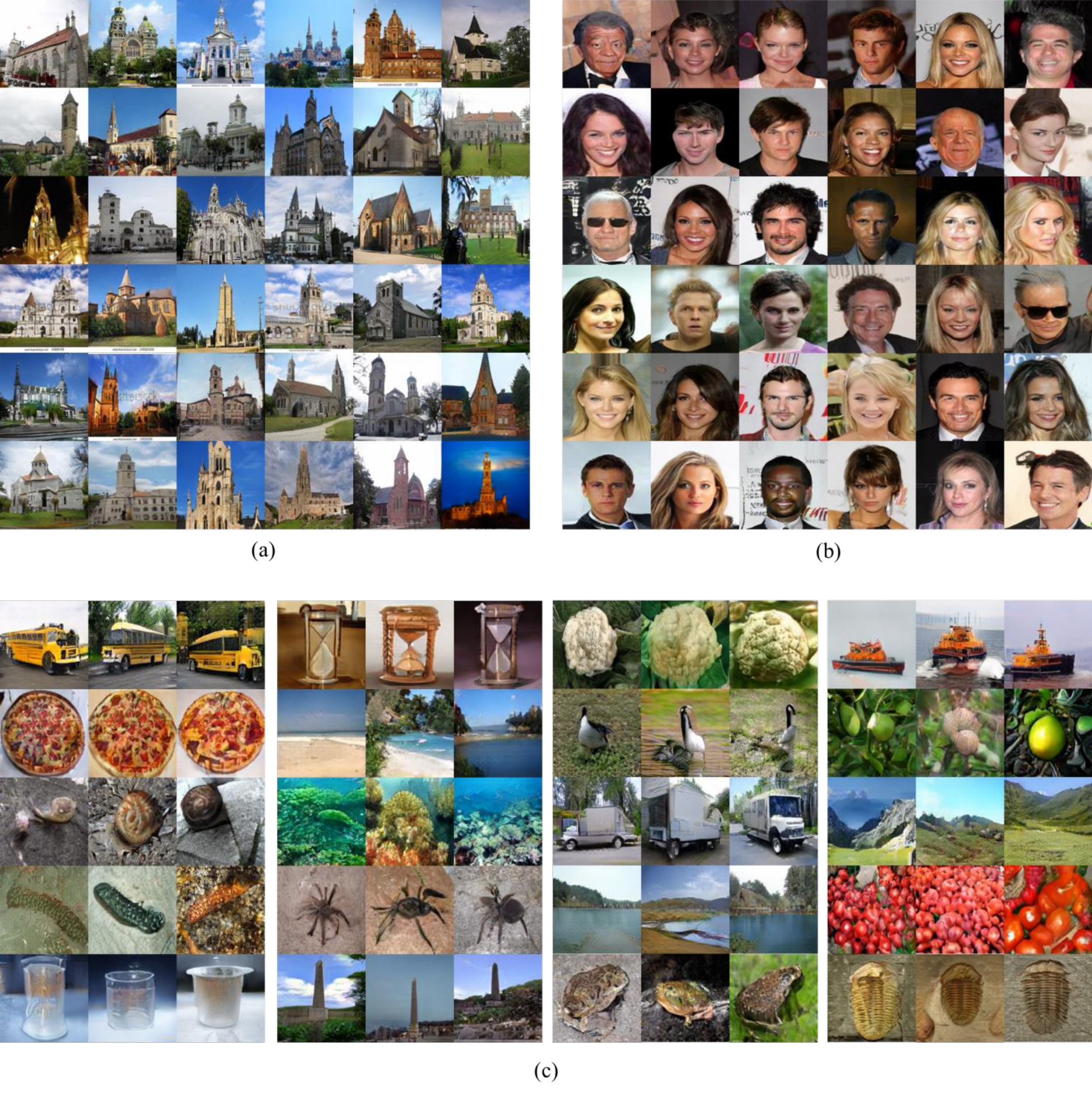}
\caption{Samples of the generated images. (a) Generated images on the LC dataset, (b) Generated images on the CA dataset, (c) Generated images on the TIN dataset.}
\label{fig:fig6}
\vspace{-0cm}
\end{figure*}

In order to validate the ability of GConv when synthesizing images in challenging datasets, we trained the GAN using the LC, CA, and TIN datasets. Specifically, the LC and CA datasets are trained by following the experimental settings for GAN, whereas the TIN dataset is trained by following that for the cGAN. In Tables~\ref{table:table2},\ref{table:table_AC}, and \ref{table:table_loss}, we already proved that the performance improvement is not due to the lucky weight initialization. Thus, we trained the network a single time from scratch. Instead, to clearly show the superiority of GConv, we drew a graph representing the FID and IS scores over training iterations, as shown in Fig.~\ref{fig:fig5}. The FID and IS scores of Conv are converged earlier than those of GConv; the proposed method shows better performance than Conv. We summarized the final FID and IS scores of GConv and those of Conv in Table.~\ref{table:table4}. The scores in the bracket indicate the minimum FID and the maximum IS scores during the training procedure. As described in Table~\ref{table:table4}, even when comparing the minimum FID score, GConv exhibits better performance than Conv. In particular, the proposed method significantly improves the GAN performance on the challenging dataset, \ie tiny-ImageNet. That means GConv is effective to produce images in various datasets. Fig.~\ref{fig:fig6} shows the samples of the generated images. As shown in Fig.~\ref{fig:fig6}, the proposed method effectively leads the generators to produce high-quality images.  

\begin{table}[t]
\caption{Comparison of GConv with Conv on the LC, CA, and TIN datasets in terms of FID and IS.}
\begin{center}
\begin{tabular}{c | c | c | c }
\hline
Dataset & Metric & Conv & GConv \\
\hline
LC & FID$\downarrow$ & 9.10 (8.88) & \textbf{7.52 (7.25)} \\
\hline
CA & FID$\downarrow$ & 9.60 (9.08) & \textbf{6.03 (6.03)} \\
\hline
\multirow{2}*{TIN} & IS$\uparrow$ & 20.23 (20.23) & \textbf{22.64 (22.64)} \\
\cline{2-4}
 & FID$\downarrow$ & 35.13 (35.13) & \textbf{29.76 (29.76)} \\

\hline
\end{tabular}
\end{center}
\label{table:table4}
\end{table}

\begin{figure*}
\centering
\includegraphics[width=1.0\linewidth]{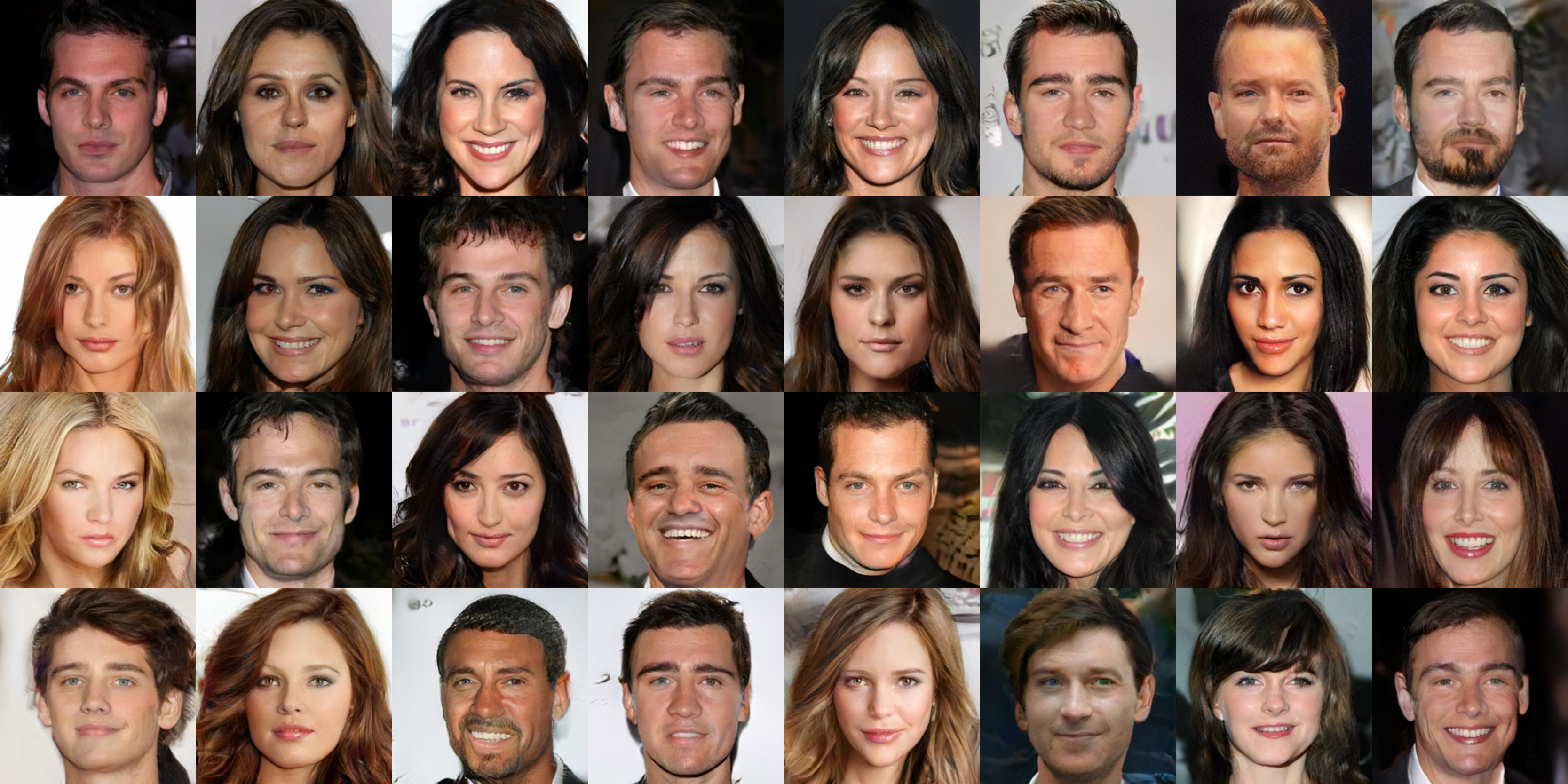}
\caption{Generated images with $256\times256$ resolutions on the Celeb-HQ dataset.}
\label{fig:fig_CAHQ}
\vspace{-0cm}
\end{figure*}

Moreover, to show the effectiveness of GConv for generating high-resolution images, we conducted additional experiments using the CelebA-HQ dataset~\cite{liu2015deep, karras2017progressive}. In this paper, we trained the networks to produce $256\times256$ images. The experimental results are presented in Table~\ref{table:table_CAHQ} and Fig.~\ref{fig:fig_CAHQ}. It is worth noting that the GConv shows much lower FID scores compared with Conv, and it produces the convincing images. Based on these results, we expected that the proposed method can utilized to produce high-resolution images. Indeed, this study does not intend to produce the design of an optimal generator and discriminator architectures for GConv. There might be another network architecture that improves the GAN performance and produces more high-quality images. Instead, this paper focuses on verifying whether it is possible to achieve the better GAN performance by simply replacing Conv with GConv.

\section{Conclusion and Future Works}
\label{sec5}
This paper has proposed a novel convolution technique for the generator, called GConv, which significantly improves the GAN performance. By simply replacing Conv with GConv, the generator is able to synthesize the visually plausible image, which results in the performance improvement of GAN. The main advantage of GConv is that it can be integrated into any generator architectures seamlessly with marginal overheads, while considerably improving the performance. In addition, this paper deeply investigated the effectiveness of GConv in various aspects by conducting many-sided ablation studies. Thus, we expect that the proposed can be applied to various GAN-based image synthesizing techniques. However, this paper aims at introducing a novel convolution method specialized to the GAN. We agree that the GConv might work well for the GAN-based image generation technique, but might cause some issues in other networks used for different applications. Although this paper only covers the effectiveness of GConv in the field of GAN-based image generation technique, we have shown that the proposed method could improve the GAN performance significantly with various aspects. As our future work, we plan to further investigate a novel convolution method that could cover various applications.

\begin{table}[t]
\caption{Comparison of the GConv with Conv on the CelebA-HQ dataset in terms of FID.}
\begin{center}
\begin{tabular}{c | c | c | c}
\hline\hline
Resolution& Metric & Conv & GConv \\
\hline
$256\times256$ & FID$\downarrow$ & 24.10 & \textbf{18.68} \\
\hline\hline
\end{tabular}
\end{center}
\label{table:table_CAHQ}
\end{table}

\section*{Declaration of Competing Interest}
The authors declare no conflict of interest.


\printcredits

\bibliographystyle{cas-model2-names}

\bibliography{egbib.bib}

\bio{}
\endbio

\endbio

\end{document}